\documentclass[12pt]{article}
\usepackage{enumerate}
\usepackage{natbib}
\usepackage{url} 
\usepackage{amsmath,amssymb,amsthm}
\usepackage{xcolor} 
\definecolor{myblue}{RGB}{0, 0, 188}
\definecolor{mygreen}{RGB}{0, 188, 0}
\usepackage[colorlinks=true, linkcolor=myblue, citecolor=mygreen]{hyperref} 
\usepackage{algorithm,algorithmic}
\usepackage{graphicx}

\newcommand{\blind}{1}

\newtheorem{theorem}{Theorem}

\newtheorem{remark}{Remark}

\newtheorem{assumption}{Assumption}

\begin{document}

\def\spacingset#1{\renewcommand{\baselinestretch}%
{#1}\small\normalsize} \spacingset{1}


\if1\blind
{
  \title{\bf Covariate-Elaborated Robust Partial Information Transfer with Conditional Spike-and-Slab Prior}
  \author{Ruqian Zhang$^{1}$\footnote{
    These authors contributed equally to this work.}, Yijiao Zhang$^{2*}$, Annie Qu$^{3}$, Zhongyi Zhu$^{1}$, Juan Shen$^{1}$\footnote{Corresponding author. Email address: shenjuan@fudan.edu.cn} \thanks{
    This work is supported by the National Natural Science Foundation of China grants 11871165, 12071087, and
12331009.}\hspace{.2cm}\\
    $^{1}$Department of Statistics and Data Science, 
       Fudan University\\
    $^{2}$Department of Biostatistics, Epidemiology and Informatics,\\ 
       University of Pennsylvania\\
    $^{3}$Department of Statistics and Applied Probability,\\ University of California, Santa Barbara}
    \date{}
  \maketitle
} \fi

\if0\blind
{
  \bigskip
  \bigskip
  \bigskip
\begin{center}
    \setlength{\baselineskip}{1.5\baselineskip}
    {\LARGE\bf  Covariate-Elaborated Robust Partial Information Transfer with Conditional Spike-and-Slab Prior}
\end{center}
  \medskip
} \fi

\bigskip
\begin{abstract}
The popularity of transfer learning stems from the fact that it can borrow information from useful auxiliary datasets. Existing statistical transfer learning methods usually adopt a global similarity measure between the source data and the target data, which may lead to inefficiency when only partial information is shared. In this paper, we propose a novel Bayesian transfer learning method named ``CONCERT'' to allow robust partial information transfer for high-dimensional data analysis. A conditional spike-and-slab prior is introduced in the joint distribution of target and source parameters for information transfer. By incorporating covariate-specific priors, we can characterize partial similarities and integrate source information collaboratively to improve the performance on the target. In contrast to existing work, the CONCERT is a one-step procedure which achieves variable selection and information transfer simultaneously. We establish variable selection consistency, as well as estimation and prediction error bounds for CONCERT. Our theory demonstrates the covariate-specific benefit of transfer learning. To ensure the scalability of the algorithm, we adopt the variational Bayes framework to facilitate implementation. Extensive experiments and two real data applications showcase the validity and advantages of CONCERT over existing cutting-edge transfer learning methods.
\end{abstract}

\noindent%
\vfill

\newpage
\spacingset{1.9} 
\section{Introduction}
\label{sec:intro}
The growing emphasis on data heterogeneity in the era of big data has brought high demand for developing efficient and robust methods to integrate information from heterogeneous data sources. The homogeneous data assumption commonly assumed in the traditional machine learning and statistical literature can lead to biased inference. Alternatively, transfer learning \citep{pan2009survey} has emerged as a powerful tool to borrow information from similar but different tasks, which can enhance the estimation efficiency on the target data. This is especially useful for high-dimensional data analysis, where the number of subjects in the target dataset could be much smaller than the number of covariates. There is a rapidly growing literature on the application of transfer learning methods in domain science, including human activity recognition  \citep{hirooka2022ensembled}, protein representation learning \citep{fenoy2022transfer}, and RNA modification identification \citep{wu2024transfer}.

One of the challenges in transfer learning lies in how to extract useful information from source datasets, and statistical research on transfer learning has grown rapidly in recent years. Among them, \citet{li2022trans}, \citet{tian2022transfer} and \citet{li2024estimation} focused on (generalized) linear models, \citet{li2023transfer} investigated Gaussian graphical models, \citet{zhang2022transfer} studied quantile models, and \citet{lin2022transfer} explored functional data. The notion of these methods is to measure the similarity between the source and target based on the global difference between parameters and to choose informative sources with small global differences adaptively. However, the global similarity measure may lead to an all-in-or-all-out principle where each source dataset is either fully utilized for information transfer or completely disregarded, resulting in inefficiency.

\begin{figure}[h]
	\centering
	\includegraphics[width=0.8\textwidth]{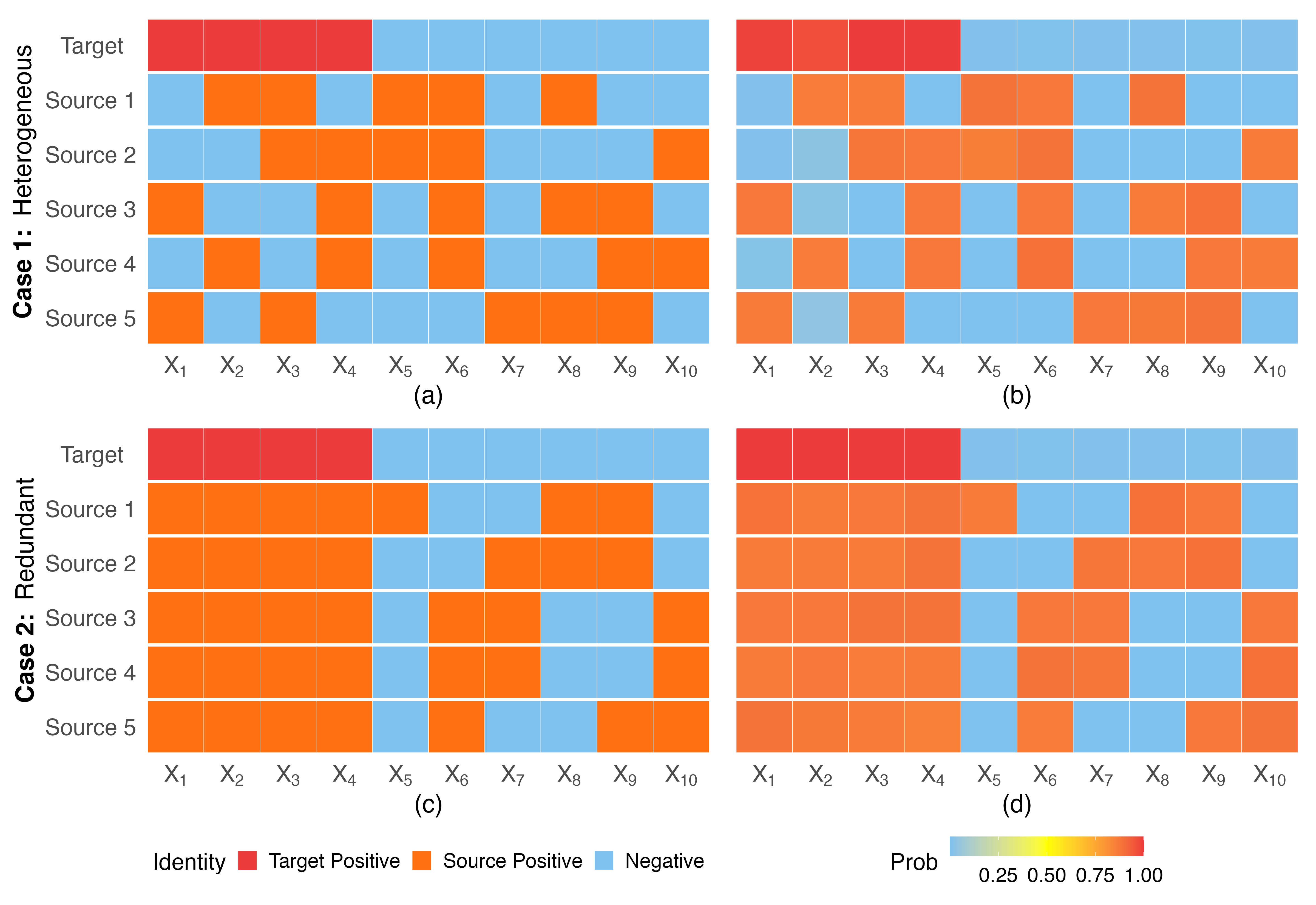}
	\caption{Illustration of the two cases of partial information similarity.
    In (a) and (c), the true variable identities are clarified with \emph{Target Positive}, \emph{Source Positive}, \emph{Target Negative}, and \emph{Source Negative} representing influential variables in the target, transferable variables in sources, non-influential variables in the target, and non-transferable variables in sources, respectively.
    In (b) and (d), the corresponding estimated posterior probabilities of all variables are shown in terms of significance in the target and transferability in sources.
    }
	\label{fig:demo_both}
\end{figure}

To understand when global similarity may fail, we illustrate two cases of partial similarity structure in the left panel of Figure \ref{fig:demo_both}. Figure \ref{fig:demo_both}(a) represents the case where sources contain \textit{heterogeneous information} and only share part of the parameters with the target, and Figure \ref{fig:demo_both}(c) demonstrates the case where sources contain all significant signals of the target but have \textit{redundant information} on additional dimensions.
Both cases are common in image classification or medical data. However, due to the global similarity requirement, existing methods may fail to recognize the sources as informative auxiliary data even if they share some partial structure. This motivates us to develop new similarity measures to fully utilize the information from sources in the absence of global similarity.

Another challenge arises from the unknown similarities between the target and source datasets. To avoid negative transfer \citep{zhang2020survey}, a two-stage framework has been developed in the transfer learning literature \citep{li2022trans, tian2022transfer, li2024estimation}, where the global similarity between sources and the target is evaluated first. Informative sources are then selected based on case-specific criteria, or a model averaging procedure is implemented. Since the underlying similarity is not known in advance, selection methods usually rely on a global similarity gap assumption between informative and non-informative sources \citep{tian2022transfer, zhang2022transfer} to ensure consistent source selection. In the absence of global similarity,  detecting similarity becomes more challenging. Therefore, a partially adaptive and robust transfer procedure is needed.

Motivated by the above challenges, we propose a novel Bayesian transfer learning method designed to enable robust partial information transfer for high-dimensional generalized linear regression. We introduce a new conditional spike-and-slab prior in conjunction with the traditional spike-and-slab prior \citep{george1993variable, Ishwaran2005ss, Rockova2018ss} for the joint distribution of parameters across both the target and sources. To characterize the partial similarity, we equip the coefficient of each covariate from each source with a conditional spike-and-slab prior centered at the corresponding target parameter.
The spike-and-slab prior can be viewed as a Bayesian approximation to the $L_0$ penalty \citep{narisetty2014bayesian}, which enables partial information transfer by preserving large discrepancies between source and target domains. This avoids the over-shrinkage issue associated with the commonly used $L_1$ penalty, which may lead to negative transfer.
We name our proposed method CONCERT for \underline{CON}ditional spike-and-slab for \underline{C}ovariate-\underline{E}laborated \underline{R}obust \underline{T}ransfer, which transfers information from informative covariates across different sources, ensuring robust integration to improve the target.
In Figure \ref{fig:demo_both}(b) and \ref{fig:demo_both}(d), we show the validity of the proposed method on joint variable and similarity selection. Details can be found in Section D.7.1 of the Supplementary Material.

Several independent lines of work provide reasons for Bayesian transfer learning. Among them, the prior distribution for the target model is assigned conditionally based on source knowledge, either directly on source data \citep{karbalayghareh2018optimal} or on source predictive density \citep{jirsa2021bayesian,papevz2022transferring}, and the conditional posterior of target response is under concern. Our method differs from these approaches by introducing a joint association of the parameters rather than directly assuming a joint distribution on the target and source data. The proposed method can significantly reduce model complexity while enhancing both efficiency and interpretability.

The CONCERT brings the following contributions to the transfer learning literature. Firstly, we allow partial information transfer. By characterizing the covariate-specific similarity for each source using the conditional spike-and-slab prior, we can fully utilize the partially useful information from different sources even in the absence of global similarity. 
We showcase the advantage over existing transfer learning methods through extensive numerical experiments and real data analysis.
Secondly, we establish the variable selection consistency. Specifically, we relax the minimal signal strength condition in the traditional high-dimensional variable selection literature in a covariate-specific manner. Our theory demonstrates that the more similar the sources are to the target on a particular variable, the lower the minimum signal strength required to detect that variable in the target. In other words, we can detect weaker signals in the target with the benefit of borrowing partial information from more sources. The estimation and prediction errors of the CONCERT also enjoy faster convergence rates than a single-task Bayesian regression.

Thirdly, the adoption of spike-and-slab prior for transfer learning provides a new perspective beyond the scope of conventional spike-and-slab prior, which is initially designed for variable selection. Under a mild local similarity gap condition, we establish the consistency of source similarity detection under the conditional spike-and-slab prior. Lastly, we utilize variational Bayes to achieve scalability with a negligible loss of accuracy. Beyond the common mean-field variational family, a novel hierarchical structure is proposed to incorporate parameter dependence. Our theoretical analysis demonstrates that the VB posterior attains near-optimal contraction rates, comparable to those of the true posterior.

The rest of the paper is organized as follows. In Section~\ref{sec:meth}, we introduce the problem setup and propose the CONCERT method for partial information transfer. A scalable variational Bayes algorithm is presented in Section~\ref{sec:algo}. Theoretical guarantees for our method are provided in Section~\ref{sec:theory}. Extensive experiments and two real data applications are provided in Sections~\ref{sec:simu} and \ref{sec:realdata}, respectively. We conclude this paper in Section~\ref{sec:conc}.

\section{Methodology}
\label{sec:meth}

In this section, we introduce our CONCERT method. Consider the high-dimensional generalized linear models with response $y\in\mathbb{R}$ and covariates $x\in\mathbb{R}^{p}$. The conditional distribution of $y$ given $x$ follows
	\begin{equation}
		\label{eq:GLM}
		p(y\mid x)= \varphi(y) \exp\{yx^{\top}\boldsymbol\beta-\psi(x^{\top}\boldsymbol\beta) \},
	\end{equation}
	where $\boldsymbol\beta\in\mathbb{R}^p$ is the parameter, and $\varphi$ and $\psi$ are known functions with $\psi^\prime(x^{\top}\boldsymbol\beta)=\mathbb{E}[y\mid x]$. Note that model (\ref{eq:GLM}) with $\psi(u)=u^2/2$ corresponds to the Gaussian linear regression model, while $\psi(u)=\log(1+e^u)$ corresponds to the logistic regression model.

Suppose we are interested in the target data ${\cal D}^{(0)}=\{(x^{(0)}_{i}, y^{(0)}_i) \}^{n_0}_{i=1}$,  where the conditional distribution satisfies (\ref{eq:GLM}):
\begin{equation*}
	p( y_{i}^{(0)}\mid x^{(0)}_i)=\varphi(y_{i}^{(0)}) \exp\{y_{i}^{(0)}x^{(0)\top}_i\boldsymbol\beta^{(0)}-\psi(x^{(0)\top}_i\boldsymbol\beta^{(0)}) \}.
\end{equation*}
However, the sample size of the target data could be insufficient under the high-dimensional setting where $n_0\ll p$. To this end, we can utilize additional auxiliary datasets. Suppose there are $K$ additional independent source datasets ${\cal D}^{(k)}=\{(x^{(k)}_{i}, y^{(k)}_i) \}^{n_k}_{i=1}$ satisfying
\begin{equation*}
	p( y_{i}^{(k)}\mid x^{(k)}_i)=\varphi(y^{(k)}_i) \exp\{y_{i}^{(k)}x^{(k)\top}_i\boldsymbol\beta^{(k)}-\psi(x^{(k)\top}_i\boldsymbol\beta^{(k)}) \},
\end{equation*}
for $k=1,\ldots, K$, respectively. 
Although additional information is available from the samples of these source datasets, the heterogeneity between the sources and the target may hinder the performance of single-task methods based solely on the target data. Our aim is to extract as much useful information as possible from the source data to help improve the performance on the target while preventing negative transfer.

\subsection{Spike-and-Slab for Variable Selection}
Before we introduce our tailored transfer learning framework, we first provide more background on the spike-and-slab prior. In the Bayesian framework, the spike-and-slab prior has been widely adopted to perform variable selection, where a mixture of distributions is specified as the prior for the parameters of interest.
Its selection properties closely resemble those of the
$L_0$ penalty \citep{narisetty2014bayesian}, making it highly effective for variable selection in high-dimensional settings.
A crucial advantage of spike-and-slab is its availability to quantify uncertainty for both parameter estimation and model selection \citep{Narisetty2019ss,Nie2022ss}.  Various mixture strategies have been adopted, including the absolute continuous spike-and-slab \citep{Biswas2022ScalableS} and the point mass spike \citep{Zeng2022ss}. In this paper, we choose the Gaussian distribution as the slab prior and the Dirac mass at zero as the spike prior. If we only had the target data, for each component $\beta^{(0)}_j$ of $\boldsymbol{\beta}^{(0)}$ for $j=1,\ldots, p$, let the prior be,
	\begin{equation*}
		\begin{aligned}
			\beta^{(0)}_{j}\mid Z_j\stackrel{ind}{\sim} Z_j N(0,\eta^2)+(1-Z_j)\delta_0,\quad
			Z_j \stackrel{iid}{\sim} \operatorname{Bernoulli}(q_0),
		\end{aligned}
\end{equation*}
where $Z_j$'s are latent variable selection indicators showing whether the $j\text{th}$ covariate of $x^{(0)}$ is informative, i.e., whether its corresponding coefficient $\beta^{(0)}_{j}$ is away from zero, $\delta_x$ denotes the Dirac mass at $x$, and $\eta$ and $q_0$ are hyperparameters. Let $S=\{j\in[p]:Z_j= 1\}$ be the index set of the target signals. Based on the likelihood and priors, one can make inferences from the posterior distribution on both variable selection and parameter estimation.

\begin{remark}
    We choose point mass spike in our prior as it is preferred for variational Bayes implementation \citep{Carbonetto2012vb, ray2022vb}. A continuous spike-and-slab prior can also be considered with the implementation of MCMC approaches.
\end{remark}

\subsection{Conditional Spike-and-Slab for Similarity Selection}
Now we illustrate how spike-and-slab priors can be leveraged for covariate-specific information transfer from sources to the target.
A source model is considered helpful for target learning in the sense that the information carried by the covariates is similar to that in the target. A simple example is when $\boldsymbol{\beta}^{(k)}\approx \boldsymbol{\beta}^{(0)}$, meaning that the entire set of covariates is helpful and fully transferable. However, as shown in Figure \ref{fig:demo_both}, a more realistic scenario is that only a fraction of the covariates are useful, that is, $\boldsymbol{\beta}^{(k)}_{G}\approx \boldsymbol{\beta}^{(0)}_{G}$ for some subset $G\subseteq \{1,\ldots,p\}$, where $\boldsymbol{\beta}^{(k)}_{G}$ denotes the subvector of coefficients indexed by $G$, while the remaining covariates are not informative. Though such source data is not fully useful, we can still selectively exploit relevant information to improve target learning. 

Motivated by this purpose, we design a novel conditional spike-and-slab prior to select the similarity structure:
\begin{equation*}
		\begin{aligned}
			\beta^{(k)}_{j}\mid (\beta^{(0)}_j, I^{(k)}_j)\stackrel{ind}{\sim} I^{(k)}_j N(\beta^{(0)}_j,\tau_k^2) +(1-I^{(k)}_j)\delta_{\beta^{(0)}_j}, \quad
			I^{(k)}_j \stackrel{iid}{\sim} \operatorname{Bernoulli}(q_k),
		\end{aligned}
\end{equation*}
where $I^{(k)}_j$'s are the latent \textit{covariate-specific similarity indicators} showing whether the $j\text{th}$ covariate of source $k$ is transferable for $j=1,\ldots, p$ and $k=1,\ldots,K$, and $\tau_{k}$ and $q_k$ are the hyperparameters. To better interpret our conditional spike-and-slab prior, we can define $\boldsymbol{\delta}^{(k)}=\boldsymbol{\beta}^{(k)}-\boldsymbol{\beta}^{(0)}$, and for each component $\delta^{(k)}_j$ of $\boldsymbol{\delta}^{(k)}$, rewrite the prior as $$\delta^{(k)}_{j}\mid (\beta^{(0)}_j, I^{(k)}_j)\stackrel{ind}{\sim} I^{(k)}_j N(0,\tau_k^2)+(1-I^{(k)}_j)\delta_{0},$$
which inherits the spirit of mixture modeling in spike-and-slab. Notably different from the traditional spike-and-slab, the conditional spike-and-slab is designed for \emph{similarity selection}, and the indicators take the role of selecting informative covariates from different sources. If $I^{(k)}_j=0$, the prior of $\beta^{(k)}_{j}$ concentrates on the point mass at $\beta^{(0)}_{j}$, showing a prior belief in $\beta^{(k)}_{j}=\beta^{(0)}_{j}$, and thus the corresponding covariate being transferable. Otherwise, if $I^{(k)}_j=1$, the prior of $\beta^{(k)}_{j}$ is away from $\beta^{(0)}_{j}$ with a large variance $\tau^2_k$, and the corresponding covariate is not considered useful for the target data. We denote $T_k = \{j \in [p]: I_j^{(k)}=1\}$ to be the index set of non-transferable signals for source $k$, and $T=\{T_k\}_{k=1}^{K}$.
The covariate-specific indicators enable us to explore partial similarity between sources and the target, addressing the limitation of existing transfer learning methods \citep{li2022trans,tian2022transfer,li2024estimation} that rely on global similarity measures such as \(\|\boldsymbol\beta^{(k)} - \boldsymbol\beta^{(0)}\|_1\) and may overlook partially relevant sources.

Given that the spike-and-slab prior provides an approximation to the $L_0$ penalty, our method can be seen as replacing the $L_1$ penalty on $\boldsymbol{\delta}^{(k)}$ in the previous transfer learning methods with an $L_0$ penalty, which allows a shift from global information transfer to partial information transfer. Specifically, if certain components of \(\boldsymbol{\delta}^{(k)}\) in source $k$ are large, indicating substantial differences between source $k$ and the target in certain dimensions, the $L_1$ penalty tends to overly suppress these differences, potentially neglecting the heterogeneity and leading to negative transfer.
As a result, existing transfer learning methods often exclude such source datasets from the transferable set in advance, limiting their contribution to target learning.

In contrast, the $L_0$ penalty mitigates excessive shrinkage of substantial discrepancies between source and target coefficients, preserving strong heterogeneity signals instead of compressing them toward zero. This allows meaningful differences between source and target domains to be retained, enabling selective information transfer from other relevant components within the same source dataset rather than discarding it entirely.

\begin{remark}
    In contrast to previous works in Bayesian transfer learning \citep{karbalayghareh2018optimal,papevz2022transferring}, our method does not impose additional assumptions on the prior distribution for the target model. Our prior for target does not involve source information and thus remains in the original form of the sole target model, which blesses it with modeling consistency and interpretability despite the existence of sources.
\end{remark}

\section{Variational Bayes Implementation}
\label{sec:algo}
The proposed Bayesian transfer learning can be carried out using MCMC sampling from full conditional posterior distributions. However, here we adopt a scalable approach based on variational Bayes (VB), which serves as a popular tool to alleviate the computational burden of MCMC in Bayesian inference \citep{Jaakkola1997,blei2017review}. The thrust of VB is to approximate the joint posterior distribution by the closest distribution within some tractable distribution family. VB is commonly used to assist stochastic search variable selection for spike-and-slab prior \citep{Menacher2023vb}, where its performance is guaranteed both empirically and theoretically \citep{Carbonetto2012vb,ray2022vb}. In this section, we derive the VB algorithm for transfer learning.

We denote the likelihood of the target data and the source data as $L({\cal D}^{(0)})$ and $L({\cal D}^{(k)})$  for $k=1,\ldots, K$, respectively. Let $\boldsymbol{\theta}^{(0)}$ and $\boldsymbol{\theta}^{(k)}$ for $k=1,\ldots,K$ denote all parameters in the target and source models, with corresponding priors $\Pi(\boldsymbol{\theta}^{(0)})$ and $\Pi(\boldsymbol{\theta}^{(k)})$. The joint prior distribution is then given by $\Pi({\boldsymbol{\theta}})=\bigotimes^{K}_{k=0}\Pi(\boldsymbol{\theta}^{(k)})$ with $\boldsymbol{\theta}=(\boldsymbol{\theta}^{(k)})^{K}_{k=0}$. Let $\cal{D}$ denote all data. By Bayes' formula, the joint posterior density can be expressed as
	\begin{equation}
		\label{eq:joint_post}
		d\Pi(\boldsymbol{\theta}\mid {\cal D})\propto L({\cal D}^{(0)})\prod^K_{k=1}L({\cal D}^{(k)})d\Pi(\boldsymbol{\theta}).
	\end{equation}
To preserve the spike‐and‐slab structure assumed in the priors, we propose a hierarchical mean-field variational family $\cal Q$, where the source coefficients remain conditional on the target coefficients, extending the framework in \citet{ray2022vb} to our transfer learning setting.
Specifically, in the target model, the variational distributions for $\beta^{(0)}_j$'s follow independent spike-and-slab distributions for $j=1,\ldots, p$:
\begin{equation*}
	\begin{aligned}
		\beta^{(0)}_j\mid Z_j \sim Z_j N(\mu^{(0)}_j, (\sigma^{(0)}_j)^{2}) + (1-Z_j)\delta_0,\quad
		Z_j \sim \operatorname{Bernoulli}(\gamma^{(0)}_j).
	\end{aligned}
\end{equation*}
In addition, for each source $k=1,\ldots,K$, the variational distributions of $\beta^{(k)}_j$'s can be represented in a conditional spike‐and‐slab structure as:
\begin{equation*}
	\begin{aligned}
		\beta^{(k)}_{j}\mid \beta^{(0)}_j, I^{(k)}_j \sim I^{(k)}_j N(\mu^{(k)}_j, (\sigma^{(k)}_j)^{2})+(1-I^{(k)}_j)\delta_{\beta^{(0)}_j},\quad
		I^{(k)}_j \sim \operatorname{Bernoulli}(\gamma^{(k)}_j).
	\end{aligned}
\end{equation*}
For other parameters $\phi^{(k)}$ specified by the specific model, we denote the variational distributions as $Q(\phi^{(k)})$, for $k=0,\ldots,K$. For example, in the linear case, $\phi^{(k)}$ represents the noise variance $(\sigma^{(k)}_y)^2$, while in the logistic case, $\phi^{(k)}$ comprises the augmented $\{\omega^{(k)}_i\}^{n_k}_{i=1}$, as detailed in Section~\ref{sec:VB_logistic}.
Combining these variational distributions, the variational family $\cal Q$ can be written in a compact and unified form for the GLM transfer learning setting as
\begin{equation}
\label{eq:vb_family}
		\mathcal{Q}=\left\{Q(\boldsymbol{\theta})=\bigotimes_{k=0}^K\left[\bigotimes_{j=1}^p\left(\gamma^{(k)}_j N(\mu^{(k)}_j, (\sigma^{(k)}_j)^2)+(1-\gamma^{(k)}_j) \Delta^{(k)}_j\right)\times Q(\phi^{(k)})\right]\right\},
\end{equation}
where $\Delta^{(k)}_j$ denotes the corresponding Dirac mass distributions with $\Delta^{(0)}_j=\delta_0(\beta^{(0)}_j)$ and $\Delta^{(k)}_j=\delta_{\beta^{(0)}_j}(\beta^{(k)}_j)$.
This factorization allows the variational family to retain the variable selection and similarity selection capabilities while ensuring tractable optimization.
The variational posterior $Q^*$ is defined as the element in $\cal Q$ that minimizes the KL divergence to the exact posterior, or equivalently maximizes the evidence lower bound:
\begin{equation}
	\label{eq:ELBO}
	Q^{*}=\operatorname{argmax}_{Q\in{\cal Q}} \mathbb{E}_Q\left[\log\frac{d\Pi(\boldsymbol{\theta}\mid{\cal D})}{dQ(\boldsymbol{\theta)}}\right].
\end{equation}
Based on $Q^*$, variable selection in the target model is determined by the variational posterior inclusion probabilities $\gamma^{(0)}_j$, where a covariate is selected if its posterior inclusion probability exceeds a prespecified threshold.
Parameter estimation is given by the variational posterior mean. Specifically, in the spike-and-slab framework, the estimated values of $\beta^{(0)}_j$ are given by $\mathbb{E}_{Q^*}[\beta^{(0)}_j]=\gamma^{(0)}_j\mu^{(0)}_j$ for $j=1,\ldots, p$.

Next, we present the primary setup of VB for linear and logistic regression, and defer explicit updates and implementation details to Section C of the Supplementary Material.

\subsection{Example: Linear Regression}
\label{sec:VB_linear}
Under the linear regression setting, the target and source models satisfy $y_{i}^{(k)}=x^{(k)\top}_i\boldsymbol\beta^{(k)}+\varepsilon_i^{(k)}$, where $\varepsilon_i^{(k)}$'s are i.i.d. random noises from $N(0,(\sigma^{(k)}_{y})^2)$ for $k=0,\ldots, K$, respectively.
We choose a commonly used inverse-Gamma prior distribution $\operatorname{I\Gamma}(a_0, b_0)$ on the additional parameters $\{(\sigma^{(k)}_y)^2\}^{K}_{k=0}$, and assume that the prior slab variances of the target and sources contain $(\sigma^{(k)}_y)^2$ in addition to $\eta$ and $\tau_k$ for conjugacy, respectively.
The hyperparameters in the weakly informative inverse-Gamma prior distributions are set as $a_0=2$ and $b_0=1$ in our numerical studies.
For the VB implementation, the variational distributions for $(\sigma^{(k)}_y)^2$ in (\ref{eq:vb_family}) are assumed to be inverse-Gamma distributions $\operatorname{I\Gamma}(a^{(k)}, b^{(k)})$ for $k=0,\ldots,K$.
	
Since the assigned prior distributions are conditionally conjugate in the linear regression scenario, coordinate ascent variational inference (CAVI) \citep{bishop2006pattern} can be used to solve (\ref{eq:ELBO}), which iteratively optimizes each factorized term while keeping the others fixed.

\subsection{Example: Logistic Regression}
\label{sec:VB_logistic}
Under the logistic regression setting, the target and source models satisfy
\begin{equation*}
    P(y_{i}^{(k)}=1\mid x_i^{(k)})=\frac{\exp\{x^{(k)\top}_i\boldsymbol\beta^{(k)}\}}{1+\exp\{x^{(k)\top}_i\boldsymbol\beta^{(k)}\}}.
\end{equation*}
Unlike linear regression models, the logistic regression models do not enjoy direct conjugacy between likelihood and spike-and-slab priors, which makes MCMC and VB schemes intractable. One way to overcome this difficulty is to adopt P\'{o}lya-Gamma (PG) data augmentation \citep{Polson2013}. Specifically, we introduce a latent PG variable $\omega^{(k)}_i$ for each binary response $y^{(k)}_i$ and therefore the augmented likelihood $L({\cal D}\mid \omega)$ is given by
\begin{equation*}
	L({\cal D}\mid \omega)=\prod^K_{k=0}\prod^{n_k}_{i=1}\frac{1}{2}\exp\left\{\left(y^{(k)}_i-\frac{1}{2}\right)x^{(k)\top}_i\boldsymbol\beta^{(k)}-\frac{1}{2}\omega^{(k)}_i(x^{(k)\top}_i\boldsymbol\beta^{(k)})^2 \right\},
\end{equation*}
where $\omega=(\omega^{(0)}_1,\ldots, \omega^{(K)}_{n_K})$ with each $\omega^{(k)}_i$ following the PG distribution $\operatorname{PG}(1,0)$ for $i=1,\ldots,n_k$ and $k=0,\ldots,K$. The corresponding variational distributions of $\omega^{(k)}_i$'s are assumed to be $\operatorname{PG}(1, c^{(k)}_i)$ with $c^{(k)}_i$ being the variational parameter. Data augmentation guarantees conjugacy so that CAVI can be utilized to solve (\ref{eq:ELBO}).

\section{Theoretical Results}
\label{sec:theory}
In this section, we study the theoretical properties of CONCERT under the linear regression setting. We first provide theoretical results for the true posterior in Section \ref{subsec:theory-tp} and further establish theoretical guarantees for the VB posterior in Section \ref{subsec:theory-vb}.

Without loss of generality, we assume that $\sigma_y^{(0)}= \cdots = \sigma_y^{(K)}$ is known in the theoretical analysis. We consider the models with $|S|\leq L_s$ and $|T_k|\leq L_t$ for some large numbers $L_s$ and $L_t$. We first introduce some notations. Let $[K]=\{1,\ldots,K\}$ and $N=\sum_{k=0}^{K}n_k$. 
Let $\boldsymbol{\theta}^{*}=(\boldsymbol{\beta}^{(0)*},\{\boldsymbol{\delta}^{(k)*}\}_{k=1}^{K})$ be the true parameters. We define the true signal set as $S^*=\{j\in[p]:|\beta_{j}^{(0)*}|\geq \zeta_{j}\}$ and the true non-transferable index set on source $k$ as $T_k^*=\{j\in[p]:|\delta_j^{(k)*}|\geq h_j^{(k)}\}$, where $\zeta_j$ is the signal level and $h_j^{(k)}$ is the heterogeneity level specified later. We use $s^*=|S^*|$ and $t_k^*=|T_k^*|$ to denote the true model size for the target and the true non-transferable model size of source $k$ for $k\in [K]$.

For a matrix $A$ and an index set $D$, let the sub-matrix of $A$ with columns in $D$ as $A_{D}$. Let $I_{d}$ be the identity matrix of order $d$. We denote the complement of a set $A$ by $A^{\mathrm{c}}$. Denote the design matrix on the $k$th dataset as $X^{(k)}$.
Then $\tilde{P}_{T_k}=X^{(k)}_{T_k}(X^{(k)\top}_{T_k}X^{(k)}_{T_k})^{-1}X^{(k)\top}_{T_k}$ and $\tilde{Q}_{T_k}=I_{n_k}-\tilde{P}_{T_k}$ are the projection matrices on the column space of $X_{T_k}^{(k)}$ and its complementary space, respectively. Let $\tilde{Q}_{T_0}=I_{n_0}$ and $\tilde{Q}$ be the diagonal block matrix of $\{\tilde{Q}_{T_k}\}_{k=0}^{K}$. We define the transformed design matrix as  $\tilde{X}^{(k)}(T_k)=\tilde{Q}_{T_k}^{1/2}X^{(k)}$ and the stacked transformed design matrix as $\tilde{\boldsymbol{X}}(T)=({X}^{(0)\top},\tilde{X}^{(1)}(T_1)^{\top},\ldots,\tilde{X}^{(K)}(T_K)^{\top})^{\top}\in\mathbb{R}^{N\times p}$. We further define $\Phi_{S,T}=\tilde{\boldsymbol{X}}_{S}(T)(\tilde{\boldsymbol{X}}_{S}(T)^{\top}\tilde{\boldsymbol{X}}_{S}(T))^{-1}\tilde{\boldsymbol{X}}_{S}(T)^{\top}$.

Given a model $\{S,T\}$, we define the \textit{covariate-$j$-specific transferable set} as $H^{(j)}_{S,T}=\{k\in[K]: j \in S\setminus T_k\}$. Denote $H_{S,T}=\cup_{j\in S}H_{S,T}^{(j)}$ as the index set of all \emph{least-transferable datasets} given $\{S,T\}$, including all sources having at least one covariate similar to the target. We further define $n^{(j)}_{S,T}=n_0+\sum_{k\in H^{(j)}_{S,T}}n_k$ to be the useful sample size for covariate $j$ and $n_{S,T}=n_0+\sum_{k\in H_{S,T}} n_k$ as the sample size of all least-informative datasets given model $\{S,T\}$. Similarly to $\tilde{\boldsymbol{X}}(T)$ defined above, we use $\tilde{\boldsymbol{X}}_{H_{S,T}}(T)$ to denote the stacked transformed design matrix consisting of all least-informative datasets in $H_{S,T}\cup\{0\}$.

\subsection{True Posterior Contraction}\label{subsec:theory-tp}
\subsubsection{Oracle theoretical results with known $T^*$}\label{subsec:theoryknown}
We first consider the oracle setting where the similarity structures $\{T^*_k\}_{k=1}^{K}$ of the sources are known in advance. For ease of notation, when there is no ambiguity, we omit the dependence of $\{{\boldsymbol{X}}(T^*), \Phi_{S,T^*}, H_{S,T^*}, n^{(j)}_{S,T^*}, n_{S,T^*}\}$ on $T^*$ in this subsection, and denote them as $\{\boldsymbol{X},\Phi_{S}, H_S,n^{(j)}_S, n_S\}$, respectively. Here are the main assumptions.

\begin{assumption}[\textbf{True model}]\label{assump:sparsesignal}
	For some large constants $C_1 , C_2>0$ and all $k\in [K]\cup\{0\}$, it holds that \noindent(i) 
	$\|{{X}}^{(k)}\boldsymbol{\beta}^{(k) *}\|_2^2 \leq C_1  n_k (\sigma_y^{(0)})^2\log p$; (ii) $\|\tilde{\boldsymbol{X}}_{S^{*c}} \boldsymbol{\beta}_{S^{*c}}^{(0) *}\|_2^2 $$\leq C_2/2 (\sigma_y^{(0)})^2 \log p$,  (iii) $\sum_{k=1}^{K}\|\tilde{X}_{T_k^{*\mathrm{c}}}^{(k)}\boldsymbol{\delta}_{T_k^{*\mathrm{c}}}^{(k)*}\|_2^2\leq C_2/2 (\sigma_y^{(0)})^2 \log p$; 
    (iv) $\min_{j\in [p]} n^{(j)}_{[p],T^*}\geq c_0 N$ for some $c_0\in(1/2,1)$.
\end{assumption}
Assumptions \ref{assump:sparsesignal}(i)-(ii) require that the order of the total signal strength of the $k$th dataset is no more than $n_k\log p$ and the non-influential signal strength in the transformed dataset is not large \citep{yang2016on}. Assumption \ref{assump:sparsesignal}(iii) imposes constraints on the magnitude of deviation within the transferable index sets across all sources. Notably, this assumption is less restrictive than the exact-transfer condition \(\boldsymbol{\delta}^{(k)*}_{T_k^{*\mathrm{c}}}=0\), as it permits the incorporation of information from approximately similar sources without requiring perfect transferability. Assumption \ref{assump:sparsesignal}(iv) requires that the informative sample size along each dimension is comparable to the total sample size, excluding true models where only a small portion of sources are useful along each dimension.

\begin{assumption}[\textbf{Design matrix}]\label{assump:desmat} \noindent(i) (Normalized columns) The design matrix on source $k$ has been normalized so that 
	$\|{{X}}^{(k)}_j\|_2^2 =n_k$ for all $j\in[p]$ and $k\in[K]\cup\{0\}$; 
	
	\noindent(ii) (Restricted eigenvalue) The gram matrix of the considered model space satisfies that $ \min_{\{S:|S|\leq L_s\}}\lambda_{\min}({\tilde{\boldsymbol{X}}_{H_S,S}^{\top}\tilde{\boldsymbol{X}}_{H_S,S}}/{n_S})\geq \lambda_l$, $\lambda_{\operatorname{min}}(X^{(0)\top}_{S^*}X^{(0)}_{S^*})\geq n_0\lambda_l$, and for all $k\in[K]$ with $|T_k^*|\geq 1$, $\lambda_{\operatorname{min}}(X^{(k)\top}_{T_k^*}X^{(k)}_{T_k^*})\geq n_k\lambda_l$, for some $\lambda_l\in(0,1)$; 
 
	\noindent(iii) (Sparse projection) Let ${Z} \sim \mathcal{N}(0, {I}_N)$. For some $C_3 \geq 8 {\lambda_l}^{-1}$, it holds that
	$$
	\mathbb{E}_Z\left[\max _{S_1,S_2:|S_1| \leq|S_2| \leq L_s} \max _{j \in S_2\setminus S_1}\frac{1}{\sqrt{n_{S_2}^{(j)}}}\left|{Z}^{T}\tilde{Q}(I_N-\Phi_{S_1})\tilde{\boldsymbol{X}}_j\right|\right] \leq \frac{\sqrt{C_3 \lambda_l \log p}}{2} .
	$$
    \noindent (iv) (Weak correlation) For all $k\in[K]$, the transformed covariates satisfy that $\|\tilde{X}_j^{(k)}\|^2_2\geq d_1n_k$ (for $j\in T_k^{*\mathrm{c}}$) and $\tilde{\boldsymbol{X}}_j^{\top}(I-\Phi_S)\tilde{\boldsymbol{X}}_j \geq d_1d_2n^{(j)}_{S}$ (for $j\in S^{\mathrm{c}}$) for all $S$ satisfying $|S|\leq L_s$,  with some $d_1, d_2\in (0,1)$.
\end{assumption}
Assumption \ref{assump:desmat} imposes conditions on the projected design, which are parallel to the assumptions for a single-task high-dimensional regression in the Bayesian literature \citep{narisetty2014bayesian,yang2016on}. Assumption \ref{assump:desmat}(ii) allows the model size $S$ to grow slowly compared to $p$ and is satisfied when $L_s\leq \min\{p,n/\log p\}$. Assumption \ref{assump:desmat}(iii) is mild and always holds under \( C_3 = O(\lambda_l^{-1}L_s) \). Assumption \ref{assump:desmat}(iv) imposes a weak dependence condition, similar to \citet{donoho2005stable} and \citet{Castillo15}. Under the special case of an orthogonal design, Assumption \ref{assump:desmat}(iv) holds for \( d_1 = d_2 = 1 \).

\begin{assumption}[\textbf{Prior}]\label{assump:prior}
	The prior satisfies (i) $q_0\sim p^{-1}$; 
	(ii) $n_{S^*} \eta^{2}\sim(n_{S^*}\vee p^{2r})$ for some $r>1+3/2C_1\lambda_1^{-1}+C_2+C_3$.
\end{assumption}
Assumption \ref{assump:prior} is on the choice of prior hyperparameters, which is similarly assumed in \citet{narisetty2014bayesian} and \citet{Narisetty2019ss}.

\begin{assumption}[\textbf{Large signal}]\label{assump:betamin} The signal strength in the target significant set $S^*$ satisfies
	\begin{equation*}
		\zeta_j^2\geq C_{\beta}\frac{(\sigma_y^{(0)})^2\log p}{n^{(j)}_{S^*,T^*}d_1d_2},  
	\end{equation*}
	for $j\in S^*$, with constant $C_{\beta}=8({3}C_1   {\lambda_l}^{-1}/2 + r + 3+(C_2+C_3)/4)$.
\end{assumption}

Assumption \ref{assump:betamin} is parallel to the $\beta$-min condition in the high-dimensional literature for variable selection consistency \citep{buhlmann2011statistics}, which requires the signal strength on the significant index set to have a large enough magnitude. Note that the minimal signal strength restriction here is covariate-specific, depending on how many informative datasets there are along with the specific dimension.

\begin{theorem}[\textbf{Model selection}]\label{thm:consistency}
	Suppose $\min_{\{S:1 \leq |S|\leq L_s\}}n_S \gtrsim \log p$. Under Assumptions \ref{assump:sparsesignal}- \ref{assump:betamin}, the posterior probability of the true model $S^*$ satisfies that,
	\begin{equation*}
		\Pi(S^*|\{y^{(k)}\}_{k=0}^{K})\geq 1-\tilde{c}_1 p ^{-1},
	\end{equation*}
    with probability at least $1-\tilde{c}_3p^{-\tilde{c}_4}$, for some constant $\tilde{c}_1,\tilde{c}_3,\tilde{c}_4>0$, where the probability is with respect to the data-generating process.
\end{theorem}

Theorem \ref{thm:consistency} can be regarded as an extension of existing results for a single-task Bayesian high-dimensional regression \citep{narisetty2014bayesian,yang2016on} to the domain of transfer learning. The proposed transfer learning framework relaxes the conditions required in the single-task setting from two perspectives.
 
First, compared to the requirements outlined in \citet{narisetty2014bayesian} and  \citet{yang2016on}, where a minimum signal strength constraint of $\min_{j\in S^*} (\beta^{(0)*}_{j})^2 \gtrsim \log p /n_0$ is necessary, we relax it in Assumption \ref{assump:betamin} by replacing the target-only sample size $n_0$ in the denominator with a covariate-$j$-specific informative size $n_{S,T^*}^{(j)}$ of all the sources and target. As the number of similar sources along the $j$th covariate becomes larger, we can detect $\beta_j^{(0)*}$ more effectively. In the ideal scenario of $j\in \cap_{k=1}^{K}T_k^{*\mathrm{c}}$, we have $n_{S^*,T^*}^{(j)}=N$, where all sources are utilized. In the worst case where all sources are non-informative on $S$, i.e., $j \in \cap_{k=1}^{K}T_k$, $n_{S^*,T^*}^{(j)}$ remains at $n_0$, resembling the case where only target data is utilized.

Second, we allow the dimension $p$ of the covariates to be higher compared to the restriction $\max\{1,s^*\}\log p \lesssim n_0$ required in \citet{yang2016on}. When all sources are least-transferable, we obtain $\min_{\{S:|S|\leq L_s\}}n_S=N$, which allows $\log p$ to go as large as the total sample size. Even in the worst case with no transferable sources, $\min_{\{S:|S|\leq L_s\}}n_S$ remains at $n_0$, showing the robustness of our method.

Based on the model selection results in Theorem~\ref{thm:consistency}, we can establish the posterior contraction rates of our proposed method in $\ell_1$/$\ell_2$ loss to guarantee parameter estimation.

\begin{theorem}[\textbf{Estimation}]
	\label{thm:estimation}
	Under the assumptions in Theorem \ref{thm:consistency} and Assumption A.0.1 in the Supplementary Material, the posterior satisfies that,
	\begin{equation}\label{eq:l1}
		\Pi\left(\boldsymbol{\beta}^{(0)}:\|\boldsymbol{\beta}^{(0)}-\boldsymbol{\beta}^{(0)*}\|_1\ge M\sigma^{(0)}_y s^*\sqrt{\frac{\log p}{n_{S^*,T^*}}}~\bigg|~\{y^{(k)}\}_{k=0}^{K}\right)\le \tilde{c}_2p^{-1},
	\end{equation}
	\begin{equation}\label{eq:l2}
		\Pi\left(\boldsymbol{\beta}^{(0)}:\|\boldsymbol{\beta}^{(0)}-\boldsymbol{\beta}^{(0)*}\|_2\ge M\sigma^{(0)}_y\sqrt{\frac{s^*\log p}{n_{S^*,T^*}}}~\bigg|~\{y^{(k)}\}_{k=0}^{K}\right)\le \tilde{c}_2p^{-1},
	\end{equation}
 with probability at least $1-\tilde{c}_3p^{-\tilde{c}_4}$, for some constant $\tilde{c}_2,\tilde{c}_3,\tilde{c}_4,M>0$.
\end{theorem}

Theorem~\ref{thm:estimation} indicates that the posterior allocates most of its mass in a neighborhood around the true parameter in terms of $\ell_1$ and $\ell_2$ loss. The size of the neighborhood depends on $n_{S^*}$, indicating that the contraction rates are faster than those when only target data is utilized, thus leading to better estimation behaviors.
The $\ell_1$ and $\ell_2$ errors converge under weaker conditions of $(s^{*})^2\log p= o(n_{S^*})$ and $s^{*}\log p= o(n_{S^*})$ than the conventional single-task 
conditions of $(s^{*})^2\log p= o(n_0)$ and $s^{*}\log p= o(n_0)$, respectively.

In existing transfer learning methods based on global similarity \citep{li2022trans,tian2022transfer}, the theoretical results require $\|\boldsymbol{\delta}^{(k)*}\|_1 \leq s^* \sqrt{\log p / n_0}$ to ensure positive transfer. As a result, source datasets with $\|\boldsymbol{\delta}^{(k)*}\|_1 > s^* \sqrt{\log p / n_0}$ are often excluded from the transferable set.
In contrast, by enabling covariate-specific information transfer, our framework allows $\boldsymbol{\delta}^{(k)*}$ to be large within $T_k^*$, thereby achieving improved error bounds when the global similarity requirement on $\|\boldsymbol{\delta}^{(k)*}\|_1$ is not satisfied.

\subsubsection{Adaptive theoretical results with unknown $T^*$}\label{subsec:theoryunknown}
In practice, the non-transferable sets $\{T^*_k\}_{k=1}^{K}$ are usually unknown. This necessitates the simultaneous detection of the significant target signal set and the source-specific non-transferable sets, which is more challenging. To achieve this, we make slight adjustments to Assumptions \ref{assump:desmat} and \ref{assump:betamin} to accommodate the case where \( T \) is unknown. Due to space limitations, we present them as Assumptions B.2.1 and B.2.2 in Section B.2.1 of the Supplementary Material. Let $C_3^{(k)}$ be some large number introduced in Assumption B.2.1. Here we present the following additional assumptions. 

\begin{assumption}[\textbf{Prior}]\label{assump:prior2} The prior satisfies (i) $q_k\sim p^{-1}$; (ii) $n_k \tau_k^{2}\sim (n_k\vee p^{2r_k})$ for some $r_k>1+r+3/2C_1\lambda_l^{-1}+C_3^{(k)}$ for $k\in [K]$.
\end{assumption}
In addition to Assumption \ref{assump:prior}, Assumption \ref{assump:prior2} imposes assumptions on the prior for the similarity of each source to the target. 

\begin{assumption}[\textbf{Large dissimilarity}]\label{assump:deltamin} We assume that the minimal dissimilarity on non-transferable set $T_k^*$ of source $k$ satisfies
	$$\min_{j\in T_k^*}(h_j^{(k)})^2\geq C_{\delta}\frac{(\sigma_y^{(0)})^2\log p}{n_kd_1d_2},$$
	for some $C_{\delta}= 8(3C_1   {{\lambda}_l}^{-1}/2+C_2/2+r_k+3+(C_2+C_3^{(k)})/4)$.
\end{assumption}
Assumption \ref{assump:deltamin} requires the minimal heterogeneity level on the non-transferable signals of source $k$ to be at least the order of $\sqrt{\log p/n_k}$ to ensure source dissimilarity detection consistency, which is in the same spirit as the minimal signal strength restriction in Assumption \ref{assump:betamin} for target signal detection consistency.

\begin{theorem}\label{thm:consistency2}
	Suppose $\min_{\{S:1 \leq |S|\leq L_s\}}n_{S,T^*} \gtrsim \log p$. Under Assumptions \ref{assump:sparsesignal}, \ref{assump:prior}, \ref{assump:prior2}, \ref{assump:deltamin}, and B.2.1-B.2.2, the posterior of $S^*, T^*$ satisfies that, 
	\begin{equation*}
		\Pi(S^*,T^*|\{y^{(k)}\}_{k=0}^{K})\geq 1-\tilde{c}_1 p ^{-1},
	\end{equation*}
  with probability at least $1-\tilde{c}_3p^{-\tilde{c}_4}$, for some constant $\tilde{c}_1,\tilde{c}_3,\tilde{c}_4>0$. Consequently, we have the covariate-specific source selection consistency, that is,
  \begin{equation*}
    \Pi(S^*,\{H_{S^*,T^*}^{(j)}\}_{j=1}^{p}|\{y^{(k)}\}_{k=0}^{K})\geq 1-\tilde{c}_1p^{-1},
  \end{equation*}
  with probability at least $1-\tilde{c}_3p^{-\tilde{c}_4}$.

  Furthermore, under Assumption A.0.1, the posterior contraction rates in terms of $\ell_1$ loss and $\ell_2$ loss in (\ref{eq:l1})-(\ref{eq:l2}) holds.
\end{theorem}
The first part of Theorem \ref{thm:consistency2} establishes both the target signal detection consistency and the covariate-specific source detection consistency simultaneously. The second part of Theorem~\ref{thm:consistency2} demonstrates that the CONCERT achieves desirable posterior contraction rates in terms of $\ell_1/\ell_2$ loss under the unknown $T^*$ case. We remark that Theorem \ref{thm:consistency2} is a non-trivial extension of the oracle results in Section~\ref{subsec:theoryknown} due to the interplay between variable selection and transferability assessment. To overcome this, we use a refined decomposition of the model space to enable a more precise and structured theoretical analysis.

\subsection{VB Posterior Contraction}\label{subsec:theory-vb}
In this subsection, we present the contraction rates for the VB posterior. Building upon the contraction results established for the true posterior, we now directly consider the case where \( T \) is unknown. To this end, we refine our assumptions regarding the prior and signal strength, as detailed in Assumptions B.3.1-B.3.3 in Section B.3.1 of the Supplementary Material. Let $\{\kappa_n\}$ be a sequence that tends to infinity at an arbitrarily slow rate, as introduced in Assumption B.3.1. We have the following results.

\begin{theorem}[\textbf{Model selection under VB posterior}]\label{thm:vb-model-selection}
    Suppose $\min_{\{S:1 \leq |S|\leq L_s\}}n_{S,T^*} \gtrsim \log p$. Under Assumptions \ref{assump:sparsesignal}, B.2.1, and B.3.1-B.3.3, the VB posterior satisfies that, 
	\begin{equation*}
		\mathbb{E}_{\boldsymbol{\theta}^{*}}Q^*(S^*,T^*)\geq 1-O(1/\kappa_n)-\tilde{c}_1p^{-1},
	\end{equation*}
  for some constant $\tilde{c}_1>0$, where  $\mathbb{E}_{\boldsymbol{\theta}^*}$ denotes the expectation with respect to the probability measure of the true data-generating process under $\boldsymbol{\theta}^*$.
\end{theorem}

Compared to the corresponding result for the true posterior in Theorem~\ref{thm:consistency2}, Theorem~\ref{thm:vb-model-selection} includes an additional $O(1/\kappa_n)$ term due to the variational approximation, implying that the VB posterior concentrates on the true model at a slightly slower rate. A similar $O(1/\kappa_n)$ term also appears in the theoretical analysis of VB by \citet{ray2022vb}.

Similarly, we can establish the VB posterior contraction rates in terms of $\ell_1$ and $\ell_2$ loss.
\begin{theorem}[\textbf{Estimation under VB posterior}]
	\label{thm:vb-estimation}
	Under the assumptions in Theorem \ref{thm:vb-model-selection} and Assumption A.0.1, the VB posterior satisfies that, 
	\begin{equation*}
		\mathbb{E}_{\boldsymbol{\theta}^*}Q^*\left(\boldsymbol{\beta}^{(0)}:\|\boldsymbol{\beta}^{(0)}-\boldsymbol{\beta}^{(0)*}\|_1\le M\kappa_n\sigma^{(0)}_y s^*\sqrt{\frac{\log p}{n_{S^*, T^*}}}\right)\geq 1-O(1/\kappa_n)-\tilde{c}_2p^{-1},
	\end{equation*}
	\begin{equation*}
		\mathbb{E}_{\boldsymbol{\theta}^*}Q^*\left(\boldsymbol{\beta}^{(0)}:\|\boldsymbol{\beta}^{(0)}-\boldsymbol{\beta}^{(0)*}\|_2\le M\kappa_n\sigma^{(0)}_y\sqrt{\frac{s^*\log p}{n_{S^*, T^*}}}\right)\geq 1-O(1/\kappa_n)-\tilde{c}_2p^{-1},
	\end{equation*}
for some constant $\tilde{c}_2,M>0$.
\end{theorem}

The results in Theorem~\ref{thm:vb-estimation} include an extra term of $O(1/\kappa_n)$ in the contraction rates and an extra factor of $\kappa_n$ in the upper bounds, compared to the true posterior counterpart in Theorem \ref{thm:consistency2}. Since \( \kappa_n \) diverges at an arbitrarily slow rate, the VB estimator converges only slightly more slowly. Similar insights have been provided in \citet{ray2020vb} and \citet{ray2022vb} for single-task Bayesian regression.

\section{Simulation Studies}
\label{sec:simu}

In this section, we empirically evaluate the performance of our proposed method, CONCERT, and compare it with various baseline target methods and state-of-the-art transfer learning methods in extensive simulation studies. Specifically, target methods include Lasso \citep{tibshirani1996}, SparseVB with Laplace slab prior \citep{ray2022vb}, and NaiveVB where CONCERT simplifies to a solo of target data. Transfer learning methods include TransLasso \citep{li2022trans} and TransGLM \citep{tian2022transfer}.

We consider both linear regression and logistic regression and present the estimation errors in three different settings. In Section~\ref{simu:setting1}, we examine the performance of all methods in the framework of informative set, a scenario commonly adopted in the transfer learning literature \citep{li2022trans}. In Section~\ref{simu:setting2}, we consider the case where sources are heterogeneous and only share partial information with the target. In Section~\ref{simu:setting3}, all sources share the whole information with the target but with redundant signals. 

Additional simulation results are provided in the Supplementary Material.
To demonstrate the robustness of our method, we present a sensitivity analysis of the spike-and-slab hyperparameters by exploring various reasonable combinations of $(q_0,\eta)$ and $(q_k, \tau_k)$ in Section D.4, where our CONCERT consistently achieves satisfactory estimation errors.
We discuss the alternative uses of hyperpriors and compare the accuracy and scalability of our VB approach with MCMC alternatives in Section D.5 and D.6, respectively.
The R implementation of our method is publicly available at \url{https://github.com/RuqianZhang/CONCERT}.


\subsection{Existence of Informative Set}
\label{simu:setting1}
Motivated by \citet{li2022trans}, we first consider the scenario where there exists a so-called informative set $\cal A$. If a source $k\in\cal A$, the similarity between this source and the target is high in the sense that the contrast $\boldsymbol\delta^{(k)}$ enjoys a small norm and sparse structure. In contrast, if a source $k\not\in{\cal A}$, i.e., is non-informative, its parameter deviates from the target parameter more severely. 

\begin{figure}[h]
	\centering
	\includegraphics[width=0.6\textwidth]{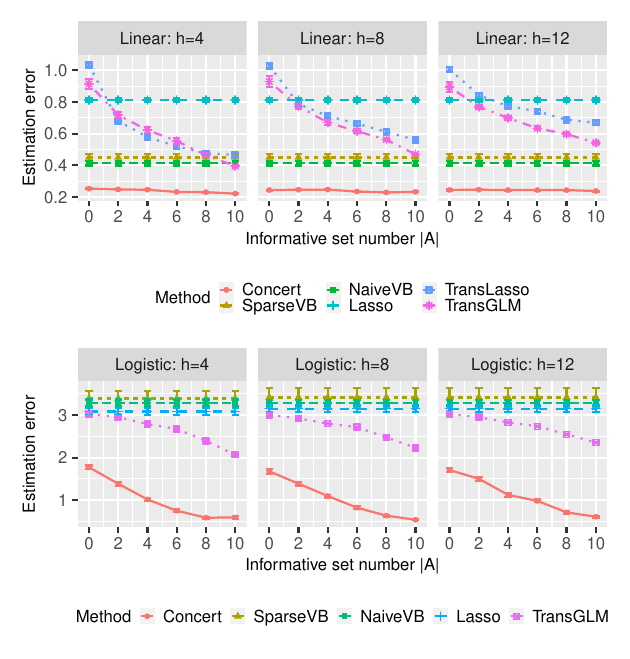}
	\caption{Estimation errors with different informative set numbers $|A|$ and varying sizes $h$.}
	\label{fig:setting1}
\end{figure}

We take $n_0=150$, $n_k=100$, for $k=1,\ldots,10$ and $p=200$. The covariates $x^{(k)}_i$ are generated independently from $\mathcal{N}(0,1)$. 
We set $\boldsymbol{\beta}^{(0)}=(\mathbf{0.5}^{\top}_{s},\mathbf{0}^{\top}_{p-s})^{\top}$ for linear regression and $\boldsymbol{\beta}^{(0)}=(\mathbf{1}^{\top}_{s},\mathbf{0}^{\top}_{p-s})^{\top}$ for logistic regression with $s=16$. Given $\cal A$, if $k\in{\cal A}$, we set
\begin{equation*}
	\beta^{(k)}_j=
	\left\{\begin{array}{ll}
		\beta_{j}^{(0)}+0.5\operatorname{Sign}(j), & j \in H_{k}; \\
		\beta_{j}^{(0)}, & \text {otherwise,}
	\end{array}\right.
\end{equation*}
where $H_k$ is a random index subset of $[p]$ with $|H_k|=h$ and $\operatorname{Sign}(j)$ takes $\pm1$ with equal probability.
If $k\not\in{\cal A}$, we set $\beta^{(k)}_j=\beta_{j}^{(0)}+\operatorname{Sign}(j)$ if $j \in H_{k}$ with $|H_k|=20$. We consider varying $|{\cal A}|$ and $h$ with $|{\cal A}|\in\{0,2,4,6,8,10 \}$ and $h\in\{4,8,12\}$. We run the experiment on 100 independent trials and summarize the means and standard errors of the root sum of squared estimation errors of $\boldsymbol{\beta}^{(0)}$ for each method. The results are reported in Figure~\ref{fig:setting1}.

As we observe from Figure \ref{fig:setting1}, CONCERT exhibits a significant advantage over all other candidates. We underscore three highlights. First, our CONCERT significantly improves the performance over single-task VB estimators, indicating the benefits of transfer learning.
Though NaiveVB methods perform worse than Lasso in the logistic setting, CONCERT outperforms TransGLM considerably.
Second, all the transfer learning methods perform worse as the size of globally informative sources decreases. However, CONCERT is much less sensitive to the decrease of $|\cal{A}|$, especially in the linear setting.
When all the sources are globally non-informative, CONCERT maintains excellent performance, while other transfer learning methods deteriorate to the single-task Lasso, or even show negative transfer. This is because the covariate-specific informative indicators $I_j^{(k)}$ in CONCERT can adaptively include locally informative covariates in different sources. Third, CONCERT is robust to the varying size $h$. As $h$ grows, other transfer learning methods show larger estimation errors, especially when $|\cal{A}|$ is large. In contrast, the CONCERT is generally unaffected by $h$ since our partial similarity modeling is able to overcome the deterioration of the globally informative sources.

\subsection{Heterogeneous Information}
\label{simu:setting2}
In this subsection, we consider a more general scenario where sources contain heterogeneous information and only share some common signals with the target. In the sense of informative set in \citet{li2022trans}, this scenario is considered non-informative.

\begin{figure}[h]
	\centering
	\includegraphics[width=0.8\textwidth]{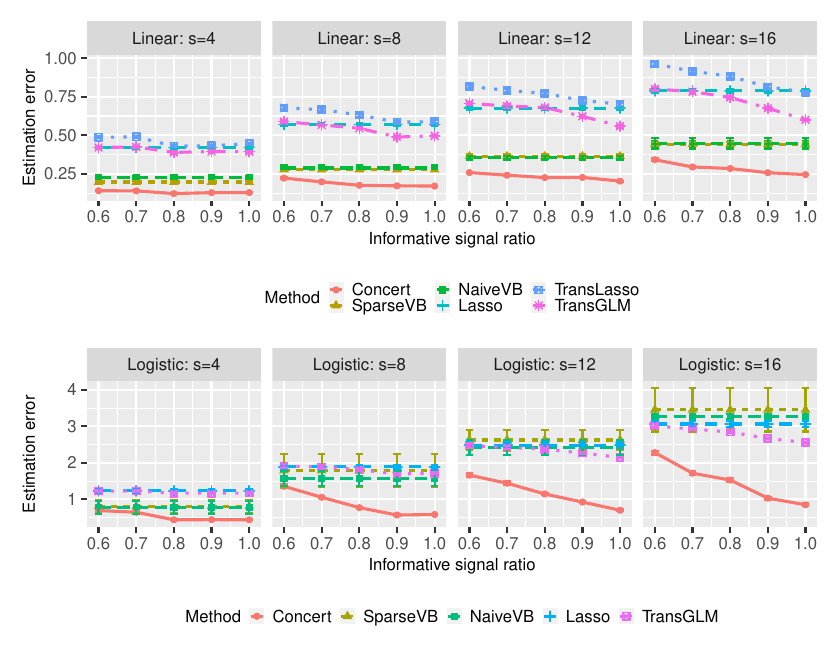}
	\caption{Estimation errors with different informative signal ratios and target signal sizes.}
	\label{fig:setting2}
\end{figure}

The covariates and noises are generated in the same manner as in Section~\ref{simu:setting1}.
For the parameters, we set $\boldsymbol{\beta}^{(0)}=(\mathbf{0.5}^{\top}_{s},\mathbf{0}^{\top}_{p-s})^{\top}$ for linear regression and $\boldsymbol{\beta}^{(0)}=(\mathbf{1}^{\top}_{s},\mathbf{0}^{\top}_{p-s})^{\top}$ for logistic regression with varying $s\in\{4,8,12,16\}$, and each $\boldsymbol{\beta}^{(k)}$ only shares partial values of $\boldsymbol{\beta}^{(0)}$. 
Specifically, we define the informative signal ratio $w$ as the ratio of shared target signals between $\boldsymbol{\beta}^{(k)}$ and  $\boldsymbol{\beta}^{(0)}$ among the first $s$ components. Then we set
\begin{equation*}
	\beta^{(k)}_j=
	\left\{\begin{array}{ll}
		0, & j \in [s]/W_{k}; \\
		\beta_{j}^{(0)}+0.5\operatorname{Sign}(j), & j \in U_{k}; \\
		\beta_{j}^{(0)}, & \text{otherwise,}
	\end{array}\right.
\end{equation*}
where $W_k$ is a random subset of $[s]$ with $|W_k|=\lceil ws\rceil$ and $U_k$ is a random subset of $\{s+1,\ldots,p\}$ with $|U_k|=20$. We set $w$ to take values in $\{0.6, 0.7, 0.8,0.9,1 \}$ and present the results based on 100 replications in Figures~\ref{fig:setting2}.

From Figure~\ref{fig:setting2}, it can be seen that the challenge of information transfer intensifies as the target information becomes more complicated and the source heterogeneity increases. However, CONCERT demonstrates significant improvement compared to NaiveVB and SparseVB. We highlight two points here. Firstly, CONCERT outperforms other transfer learning methods significantly. As all sources are globally non-informative in most cases, TransLasso and TransGLM fail to extract useful knowledge from source data, which makes them fail to compete with Lasso on target data. In contrast, CONCERT succeeds in identifying locally informative covariates in most cases and shows strong robustness when the amount of shared information is not dominant.
Secondly, though the performance of all transfer learning methods improves as the informative signal ratio becomes larger, CONCERT is much more responsive. Additionally, we note that the advantage of CONCERT over other candidates becomes more apparent as the sparsity number $s$ increases.

\subsection{Redundant Information}
\label{simu:setting3}
We further investigate the scenario where all sources contain complete target signals, however, with various degrees of contamination on the rest.
Though intrinsically these sources are all excellent candidates for transfer learning, many of them are regarded as globally non-informative and will be removed under the framework of informative set.

\begin{figure}[h]
	\centering
	\includegraphics[width=0.8\textwidth]{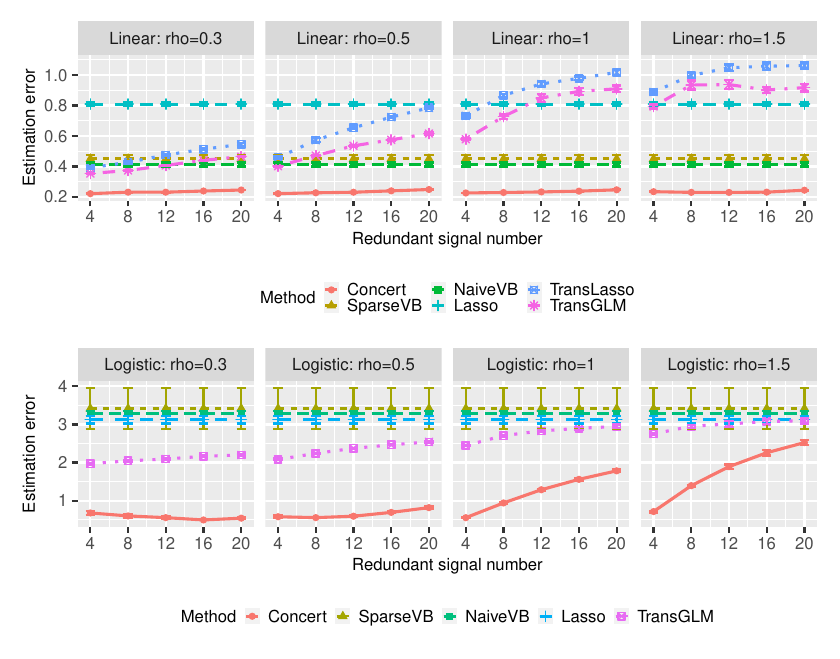}
	\caption{Estimation errors with different redundant signal numbers and strengths $\rho$.}
	\label{fig:setting3}
\end{figure}

We adopt the same data-generating process for covariates and noises as before. We set the target parameter as $\boldsymbol{\beta}^{(0)}=(\mathbf{0.5}^{\top}_{s},\mathbf{0}^{\top}_{p-s})^{\top}$ for linear regression and $\boldsymbol{\beta}^{(0)}=(\mathbf{1}^{\top}_{s},\mathbf{0}^{\top}_{p-s})^{\top}$ for logistic regression with $s=16$ and the source parameters are constructed according to
\begin{equation*}
	\beta^{(k)}_j=
	\left\{\begin{array}{ll}
		\beta_{j}^{(0)}+\rho\operatorname{Sign}(j), & j \in U_{k}; \\
		\beta_{j}^{(0)}, & \text {otherwise,}
	\end{array}\right.
\end{equation*}
where $\rho$ represents the redundant signal strength with $\rho\in\{0.3, 0.5, 1, 1.5\}$ and $|U_k|\in\{4,8,12,16,20 \}$ is the redundant signal number. We repeat the experiment 100 times, and the summarized results are shown in Figure~\ref{fig:setting3}.

Figure~\ref{fig:setting3} shows that CONCERT can avoid being misled by redundant signals and performs the best. While other transfer learning methods deteriorate rapidly as the redundant signal numbers and strengths grow, CONCERT can identify locally informative covariates, which reduces its bias even when the magnitude of redundant signals dominates that of true signals.
We highlight that other transfer learning methods may suffer from negative transfer, especially in the cases of large redundant signal strengths, while CONCERT remains robust in all cases and is generally not affected by redundant signals in linear regression.

\section{Real Data Analysis}
\label{sec:realdata}
In this section, we apply CONCERT to GTEx data and Lending Club data to demonstrate its effectiveness under the linear and logistic regression settings, respectively.

\subsection{Application to GTEx Data}
\label{sec:GTEx}
The proposed transfer learning algorithm is applied to the Genotype-Tissue Expression (GTEx) data, accessible at https://gtexportal.org/. This dataset encompasses gene expression levels from 49 tissues, involving 838 individuals and aggregating 1,207,976 observations of 38,187 genes. Building upon the methodology outlined by \citet{li2022trans}, our investigation focuses on gene regulations within the central nervous system (CNS) across diverse tissues. The collection of CNS-related genes is consolidated under MODULE\_137, comprising a total of 545 genes, alongside an additional 1,632 genes that exhibit significant enrichment in the same experiments as those within the module. 

Our specific interest lies in the prediction of the expression level of the gene SH2D2A (SH2 Domain Containing 2A) in a target tissue through the utilization of other central nervous system genes. SH2D2A encodes a T-cell-specific adapter protein that regulates early T-cell activation. It plays a crucial role in inflammatory neuropathies, particularly chronic inflammatory demyelinating polyneuropathy \citep{Uncini2011polumorphism}. Six brain tissues are considered as target tissues and, for each target, the remaining 48 tissues serve as sources. The corresponding models are individually estimated. We first screen the genes by examining their correlation with the expression level of SH2D2A, leaving 149 genes whose correlation coefficients are greater than 0.25. All the expression levels are standardized. The average sample size for the target dataset is 194, while for the sources, it is 17,135. We randomly split the data into a training set (80\% of all samples) and a testing set (20\%) 50 times, and compare the performance of CONCERT with the same candidates as listed in Section~\ref{sec:simu}. The average prediction errors on the testing sets are reported in Table~\ref{table:SH2D2A}.

\begin{table}[h]
\center
\caption{Prediction errors (standard errors) over 50 replications on 6 different brain tissues.\label{table:SH2D2A}}
\resizebox{\textwidth}{!}{\begin{tabular}
{c@{\hspace{1\tabcolsep}}c@{\hspace{1\tabcolsep}}c@{\hspace{1\tabcolsep}}c@{\hspace{1\tabcolsep}}c@{\hspace{1\tabcolsep}}c@{\hspace{1\tabcolsep}}c}
\hline
Method  & A.C.Cortex & Amygdala  & Cortex        & Hippocampus   & N.A.B.Ganglia & S.Nigra \\
\hline
\textbf{CONCERT}  & \textbf{0.214 (0.046)} & \textbf{0.151 (0.065)}  & \textbf{0.617 (0.092)} & \textbf{0.356 (0.064)} & \textbf{0.529 (0.166)} & \textbf{0.690 (0.124)}\\
SparseVB & 0.449 (0.354) & 0.232 (0.202)  & 0.830 (0.249) & 0.426 (0.143) & 0.709 (0.238) & 0.821 (0.161)\\
NaiveVB  & 0.279 (0.108) & 0.163 (0.085) & 0.820 (0.236) & 0.441 (0.155) & 0.633 (0.178) & 0.806 (0.157)\\
Lasso    & 0.544 (1.033) & 0.176 (0.134) & 0.725 (0.186) & 0.460 (0.286) & 0.613 (0.198) & 0.754 (0.176)\\
TransLasso & 0.235 (0.078) & 0.172 (0.049) & 0.679 (0.160) & 0.424 (0.210) & 0.564 (0.207) & 0.717 (0.133)\\
TransGLM & 0.245 (0.153) & 0.164 (0.075) & 0.705 (0.178) & 0.430 (0.127) & 0.565 (0.172) & 0.762 (0.141)\\
\hline
\end{tabular}}
\end{table}

In general, CONCERT performs the best among all the methods. We underscore three findings. Firstly, in all studies, CONCERT makes improvements over SparseVB and NaiveVB which solely use target information, indicating that at least one of the sources is partially informative. Secondly, CONCERT outperforms all other transfer learning methods and enjoys a lower standard deviation. Specifically, for the Hippocampus tissue, CONCERT reduces the prediction error of Lasso by more than 20\%, whereas TransLasso and TransGLM only reduce it by less than 10\%. This suggests that the source datasets for Hippocampus may have limited useful global information, but there are many local pieces of information that are valuable. Such partial information can be captured by CONCERT, thereby further reducing prediction errors. 

Lastly, we observe negative transfer in the Nigra tissue when using TransGLM. This may be due to the fact that the source detection method in TransGLM is based on the global similarity gap assumption, which may not hold here.  In contrast, the CONCERT remains robust and provides an improvement over target-only methods.

\subsection{Application to Lending Club Data}
In this subsection, we study the issuing of loan defaults using the Lending Club data available through Kaggle at https://www.kaggle.com/datasets/wordsforthewise/lending-club/. Lending Club is a financial technology company which offers peer-to-peer (P2P) lending services through its online platform, and the Lending Club dataset contains features on both borrowers and loans issued. 
We consider a binary response based on the loan status, where loans that are not fully paid are denoted as class 1, and otherwise denoted as class 0.
Our goal is to predict the risk of loan default within the target domain using information from other source domains.

We focus on the period from 2007 to 2011. After excluding the variables whose values are missing for most records and adding the pairwise interaction terms between the remaining variables, we obtain 3,458 loan records and 190 predictors in six domains: vacation, medical, wedding, education, moving, and house.
We consider all domains as our targets and estimate the corresponding models individually. For each target, all other domains are utilized as sources. 
Similarly to Section~\ref{sec:GTEx}, we randomly split the data into a training set and a testing set for 50 times, and compare the CONCERT with the same candidates for the logistic setting in Section~\ref{sec:simu}. The cutoff for binary classification is set as $0.5$, and the average prediction errors on the testing sets are reported in Table~\ref{table:LC}.

\begin{table}[h]
	\center
	\caption{Prediction errors (standard errors) over 50 replications on 6 different loan types.\label{table:LC}}
	\resizebox{\textwidth}{!}{\begin{tabular}
	{c@{\hspace{1\tabcolsep}}c@{\hspace{1\tabcolsep}}c@{\hspace{1\tabcolsep}}c@{\hspace{1\tabcolsep}}c@{\hspace{1\tabcolsep}}c@{\hspace{1\tabcolsep}}c}
	\hline
	Method     & Vacation & Medical & Wedding & Education & Moving & House\\
	\hline
	\textbf{CONCERT}    & \textbf{0.141 (0.038)} & \textbf{0.173 (0.029)} & \textbf{0.106 (0.019)} & \textbf{0.208 (0.045)} & \textbf{0.160 (0.027)} & \textbf{0.167 (0.040)}\\
	SparseVB   & 0.154 (0.045) & 0.175 (0.028) & 0.108 (0.019) & 0.212 (0.046) & 0.165 (0.027) & 0.185 (0.035)\\
	NaiveVB    & 0.151 (0.040) & 0.177 (0.028) & 0.111 (0.021) & 0.222 (0.052) & 0.167 (0.029) & 0.188 (0.037)\\
	Lasso      & 0.388 (0.127) & 0.438 (0.041) & 0.379 (0.040) & 0.361 (0.079) & 0.595 (0.156) & 0.367 (0.045)\\
	TransGLM   & 0.156 (0.039) & 0.174 (0.027) & 0.108 (0.020) & 0.213 (0.050) & 0.164 (0.028) & 0.178 (0.039)\\
	\hline
	\end{tabular}}
	\end{table}

Table \ref{table:LC} shows the advantage of the CONCERT on improving the prediction accuracy of loan default on the target domain. It is noteworthy that Bayesian methods such as SparseVB and NaiveVB can perform much better than Lasso when only the target dataset is used. Although TransGLM shows improvements over Lasso, its performance is only comparable to that of the single-task Bayesian methods. In contrast, our method can further enhance the prediction accuracy of the single-task Bayesian methods. Specifically, in the domains of vacation and house, the CONCERT reduces the prediction errors of SparseVB and NaiveVB by about 10\%.

\section{Conclusion}
\label{sec:conc}
In this paper, we propose a novel Bayesian method to allow partial information transfer. By imposing a conditional spike-and-slab prior together with the traditional spike-and-slab prior on the joint distribution of parameters in the sources and the target, we can achieve variable selection and information transfer simultaneously, robustly, and adaptively. Our theoretical results demonstrate the validity of CONCERT for robust partial information transfer. We adopt the variational Bayes framework to ensure scalability of the algorithm. Numerical experiments and real data analysis show that the CONCERT is robust to source heterogeneity and can better utilize partially informative sources to improve target learning.

There are several potential directions for future work.
First, although the simulation results show strong performance in both linear and logistic regression models, our current theoretical framework only covers the linear case. Extending the theory to logistic regression requires handling the parameter-dependent Hessian that arises from the nonlinearity of the model, which can be addressed using techniques in \citet{Narisetty2019ss}.
Second, inference procedures may be developed, which is a significant advantage of Bayesian methods. Moreover, the innovative use of the conditional spike-and-slab prior is not limited to transfer learning. This idea can be further applied to other settings where both shared components and subject-specific components exist, such as data integration or robust learning. These extensions are beyond the scope of this paper and require further development.

\paragraph{Acknowledgements:}
The authors would like to thank the reviewers, an Associate Editor and the Editor for their constructive comments that improved the quality of this paper.

\paragraph{Disclosure statement:}
The authors report there are no competing interests to declare.

\bibliographystyle{Chicago}

\bibliography{Bibliography-MM-MC}
\end{document}